\documentclass{article}
\usepackage[final]{colm2026_conference}
\usepackage{graphicx}
\usepackage{booktabs}
\usepackage{amsmath,amssymb}
\usepackage{tabularx,array}
\usepackage[hidelinks]{hyperref}
\newcolumntype{Y}{>{\raggedright\arraybackslash}X}
\newcommand{\pol}[1]{\texttt{#1}}
\newcommand{\TraceAuthors}{Radhika Gaonkar\\Prime Intellect}
\newcommand{\TraceHeader}{Published at the AI Measurement Science Workshop at COLM 2026.}

\title{TRACE: Diagnosing Verifier Brittleness in\\ Agentic Evaluation}
\author{\TraceAuthors}
\hypersetup{pdftitle={TRACE: Diagnosing Verifier Brittleness in Agentic Evaluation},pdfauthor={Radhika Gaonkar}}

\begin{document}
\maketitle
\lhead{\TraceHeader}

\begin{abstract}
Verifier scores now serve as both benchmark metrics and training rewards for large language model (LLM) agents, and a change in score is routinely read as a change in capability. It may instead reflect a change in the evaluation. We introduce TRACE, a protocol that turns a score change from a verdict into a testable diagnosis: it applies a targeted change to one part of an evaluation, compares paired runs, checks whether the agent's behavior changed, and rescores unchanged trajectories to test whether the scoring rule is responsible. In a controlled suite of 25 synthetic tasks, renaming tools lowers a scripted agent's score by 0.250 even though it performs exactly the same operations; restoring the original names at scoring time closes the entire gap, while the same mutation exposes a genuine behavioral failure in a second agent. On public $\tau^2$-bench tasks with four LLM agents, an initial 30-task study finds mixed reward changes whose one clear effect does not replicate. In a larger follow-up on 88 new tasks with repeated runs per condition, renaming tools or reformatting tool outputs leaves reward unchanged to within $\pm0.10$ for seven of eight agent--change pairs, whereas tool names that deliberately mislead lower every agent's reward by 0.20--0.44, showing that the setup can detect real effects. Identical reruns flip 15--36\% of task outcomes, so single-run comparisons cannot separate presentation effects from run-to-run variation. Two frontier LLM judges give consistent verdicts when a fixed trajectory is presented differently, yet disagree with each other on 57\% of the same records, largely because one grades procedure rather than outcome. TRACE thus separates what a score change says about the agent from what it says about the measurement.
\end{abstract}

\section{Introduction}

Large language model (LLM) agents are increasingly evaluated, compared, and trained through the scores of automated verifiers. When such a score changes, the natural reading is that the agent has become more or less capable. That reading can be wrong. In one of our controlled experiments, we rename an agent's tools without changing what they do; the agent performs exactly the same operations, yet its score drops by 0.250, because the verifier matches tool names rather than checking the operations performed. Reported on a leaderboard, this drop would record a regression that never happened; used as a training reward, it would penalize correct behavior.

Here a \emph{verifier} is any rule-based or learned evaluator that scores an agent's answer, actions, or resulting task state. Its scores increasingly act as rewards for reinforcement learning on reasoning and agentic tasks \citep{deepseek2025r1, da2025agentrlvr}. Work on specification gaming and reward-model overoptimization shows how optimizing a score can lead away from the intended goal \citep{krakovna2020specification, gao2023overoptimization, akter2026proxygaming}. Interactive benchmarks add dependencies on tool interfaces, observation formats, simulated users, and LLM judges \citep{yao2024taubench, zhou2023webarena, jimenez2023swebench}, and audits of agentic benchmarks find task and reward-design flaws that misstate capability \citep{zhu2025abc}. Following the measurement-validity view \citep{messick1995validity, bean2025measuring}, we ask what conclusions a verifier score actually supports.

The difficulty is that a score gap under a perturbation is ambiguous: the agent, the environment, or the verifier may be responsible, and the gap alone cannot tell them apart. We introduce TRACE, named for the agent traces it examines: a diagnostic protocol that resolves this ambiguity with evidence beyond the score under test (\S\ref{sec:protocol}). By comparing trajectories and rescoring unchanged ones, TRACE turns robustness testing from asking \emph{whether} a score changed into asking \emph{why}. We validate it in settings of increasing realism. In a controlled suite where the source of each failure is known, TRACE separates a tool-name scoring defect ($0.250\rightarrow0.000$ after rescoring) from genuine behavioral failures under the same mutation (\S\ref{sec:experiments}). In a frozen public $\tau^2$-bench study with four LLM agents, fresh paired runs measure whole-system sensitivity, and a fixed-trajectory judge audit isolates evaluator sensitivity (\S\ref{sec:external}). Larger follow-ups with repeat and positive controls show that its one clear effect does not replicate, that meaning-preserving changes are equivalent within $\pm0.10$ while misleading names are detected, and that consistent judges can still disagree on what counts as success (\S\ref{sec:study3}). Across all settings, the same lesson holds: a score change is an observation, not an explanation.

\section{TRACE: From Score Differences to Explanations}
\label{sec:protocol}

TRACE starts with a task, an agent, and a verifier. A \emph{mutation} is a targeted change to the evaluation; a \emph{probe} is the paired test it defines. A run produces a \emph{trajectory}: the record of actions, observations, and final answer. For task and environment $x$, policy $\pi$, seed $\xi$, and mutation $m$, we compare
\[
r^0=\operatorname{Rollout}(\pi,x,\xi),\qquad
r^m=\operatorname{Rollout}(\pi,m(x),\xi),
\]
holding the base task, policy, and seed fixed. $S(r)$ denotes the verifier score of trajectory $r$. Because some mutations change the scoring rule itself, each trajectory is scored under its own evaluation condition.

\begin{figure}[t]
\centering
\includegraphics[width=\linewidth]{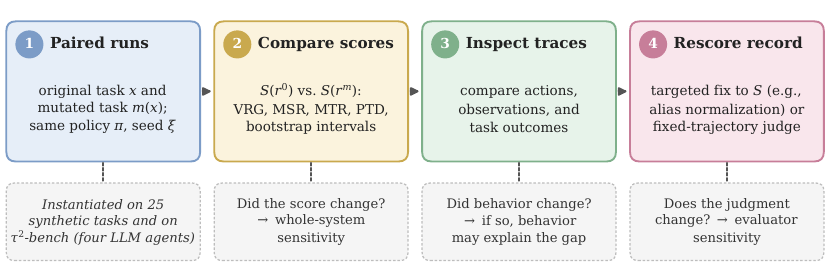}
\caption{The TRACE protocol. Paired runs (1) yield score-level diagnostics (2); trajectory inspection (3) checks whether behavior changed, and rescoring the same record under a targeted correction (4) tests the evaluator. Dashed boxes give the question each step answers.}
\label{fig:protocol}
\end{figure}

The workflow (Figure~\ref{fig:protocol}) has four steps: run paired tests, compare scores, inspect the trajectories, and test a suspected scoring error by \emph{counterfactual rescoring}, which corrects the scoring rule and rescores the same trajectory without rerunning the agent. These steps separate three questions that are often conflated: a paired rerun asks whether the \emph{system-level score} changed; trajectory comparison asks whether \emph{agent behavior or task outcome} changed; and rescoring an unchanged trajectory asks whether the \emph{evaluator} changes its judgment. We call sensitivity at the first level \emph{whole-system sensitivity} and at the third \emph{evaluator sensitivity}; only the latter is direct evidence about the verifier. A change in benchmark reward and a changed judgment on the same record are thus different findings. Two controls make a paired difference interpretable: an identical rerun estimates how often outcomes change by chance, and a \emph{positive control}, a change known to matter, shows that the pipeline can detect a real effect.

\subsection{What the Probes Change}

The six probe families of the controlled study (Table~\ref{tab:probes}) do not all predict equal scores. Tool renaming preserves tool behavior, so equivalent operations should receive equal credit; a rubric change alters what earns credit, so scores are expected to differ. Other probes change requirements, incentives, or access to information; the incentive probe targets a \emph{proxy}, a visible reward that stands in for the intended outcome. We therefore interpret each result against its intended change rather than treating every score decrease as an error. Because each family uses a single template (e.g., the wording probe prepends an instruction rather than sampling paraphrases), results characterize these specific transformations, not the full space of changes in each family.

\begin{table}[htbp]
\centering
\small
\renewcommand{\arraystretch}{1.12}
\begin{tabularx}{\linewidth}{@{}>{\raggedright\arraybackslash}p{0.18\linewidth}Y Y@{}}
\toprule
Probe & Implemented change & Interpretation \\
\midrule
Task wording & Prepend a fixed instruction to preserve constraints. & Limited wording check; not a diverse paraphrase test. \\
Tool interface & Rename tools while preserving dispatch and arguments. & Check whether equivalent operations receive equivalent credit. \\
Rubric & Change component weights and require an evidence mention. & Measure sensitivity to a different scoring rule. \\
Added constraint & Require evidence citation in the final report. & Test an additional reporting requirement. \\
Proxy incentive & Increase visible incentives on applicable reward-hacking tasks. & Compare the proxy with simulator task quality. \\
Observation access & Return partial experiment results; expose details through log inspection. & Test behavior when evidence requires an extra access step. \\
\bottomrule
\end{tabularx}
\caption{The six probe families of the controlled study, each implemented by one fixed template. Not every probe changes every task's execution; incentive changes, for example, apply only to tasks with proxy channels.}
\label{tab:probes}
\end{table}

\subsection{Score-Level Diagnostics}
\label{sec:metrics}

\emph{How much did the score change?} The \emph{verifier robustness gap} (VRG) is the original score minus the mutated score, so a positive gap means the score fell:
\[
\mathrm{VRG}(r^0,r^m)=S(r^0)-S(r^m).
\]

\emph{Does the agent still pass?} A small gap can also mean failure in both conditions. The \emph{mutated success rate} (MSR) is the fraction of mutated runs that pass the verifier, and the \emph{mutation transfer rate} (MTR) restricts this to pairs whose original run passed:
\[
\mathrm{MSR}_{\tau}=\Pr[S(r^m)\geq\tau],\qquad
\mathrm{MTR}_{\tau}=\Pr[S(r^m)\geq\tau\mid S(r^0)\geq\tau].
\]
We use pass threshold $\tau=0.65$; the artifacts also report $0.55$ and $0.75$. Both are verifier pass rates, not independent checks of task completion. MTR counts matched pairs and is undefined when no original run passes.

\emph{Does reward agree with task quality?} For a visible proxy reward $P(r)$ and a simulator-defined quality signal $H(r)$, the \emph{proxy--true divergence} is
\[
\mathrm{PTD}(r)=P(r)-H(r).
\]
A positive value means the proxy exceeds simulator quality; it does not imply intent to exploit. Scales differ across tasks and the proxy may exceed one, so we report PTD only for the 12 tasks with proxy channels (Appendix~\ref{app:scoring}).

\section{Study 1: Diagnosing Known Failures}
\label{sec:experiments}

A diagnostic is trustworthy only if it recovers causes we already know, so we first apply TRACE where failures are built in: scripted agents with prescribed behavior and a verifier with a known defect.

\paragraph{Environment and verifier.}
\texttt{ResearchOpsEnv} contains 25 synthetic research-operations tasks, such as correcting a learning rate or preventing data leakage, in which agents retrieve documents, edit configurations, run simulated experiments, inspect logs, and submit reports. The simulator records task quality $H$ and proxy rewards $P$. The composite verifier, distinct from $H$, combines report-target terms, tool-name coverage, constraint and efficiency checks, and a hack-risk penalty (Appendix~\ref{app:scoring}).

\paragraph{Agents with known behavior.}
Three scripted policies have prescribed behavior: \pol{careful} applies the intended fix and gathers evidence; \pol{overfit} targets expected answer tokens and scoring cues with limited evidence, and has an explicit branch for the tool-renaming condition; \pol{shortcut} manipulates proxy settings or submits an unsupported report. Both \pol{careful} and \pol{overfit} can read the intended fix. They are controlled examples, not models of what a learned agent would discover.

\paragraph{Paired runs and uncertainty.}
The archive holds 1,575 records: 525 task--policy--condition cells (25 tasks, seven conditions, three policies), each repeated under three seed labels. The scripted policies are deterministic, so the repeats are identical, and re-executing the archived code reproduces every record exactly; we estimate uncertainty by resampling base tasks (1,000 task-cluster bootstrap samples, seed 17), so intervals describe variation within this designed suite.

\subsection{The Same Execution Can Receive a Lower Score}
\label{sec:repair}

\begin{figure}[t]
\centering
\includegraphics[width=\linewidth]{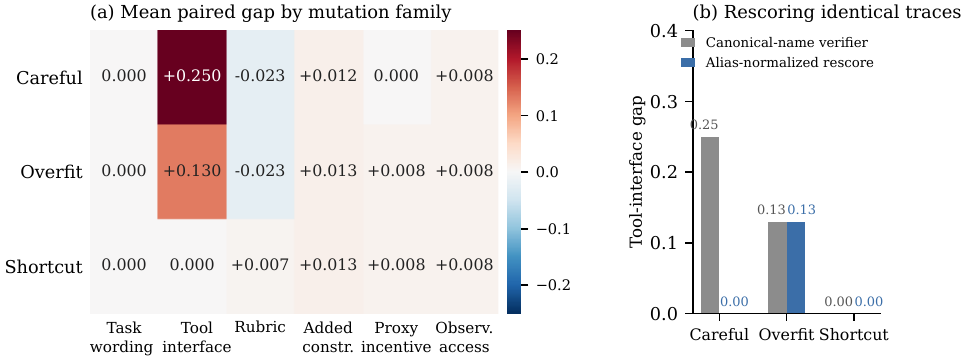}
\caption{Score gaps and their causes. (a) VRG (mean original-minus-mutated score; positive means a lower score under the mutation) by policy and probe over the 1,575 archived runs. (b) Mapping tool names back before rescoring removes \pol{careful}'s gap but not \pol{overfit}'s behavioral failure; the zero gap for \pol{shortcut} reflects poor behavior in both conditions.}
\label{fig:repair}
\end{figure}

Tool renaming produces \pol{careful}'s largest mean gap, 0.250 (Figure~\ref{fig:repair}a); its other gaps are at most 0.023 in either direction. Its paired trajectories have identical observations and simulator quality; only the verifier's name matching changes, crediting none of the five required tool names once \pol{careful} uses the advertised aliases.

To test this explanation, we map aliases back to the original names and rescore, holding the trajectories and all other reward components fixed. The gap for \pol{careful} falls from 0.250 to 0.000 (Figure~\ref{fig:repair}b), so the name-matching rule explains the entire gap for this policy and mapping.

The same correction leaves \pol{overfit}'s gap at 0.130. Under renaming, \pol{overfit} takes a different scripted branch and submits a short, unsuccessful report, which no rescoring can repair. One probe thus exposes a scoring defect for one agent and a behavioral failure for another, a distinction the gap alone cannot make.

\subsection{Small Gaps Can Hide Consistent Failure}

Gaps can also mislead in the opposite direction. The \pol{careful} policy passes most mutated runs (MSR 0.860), against 0.053 for \pol{overfit} and 0.000 for \pol{shortcut} (Table~\ref{tab:calibration}). Yet \pol{shortcut} has the smallest mean gap, 0.006, because it scores poorly in both conditions. A gap-only robustness report would rank it as the most robust policy; absolute pass rates show that its stability reflects consistent failure.

\begin{table}[htbp]
\centering
\small
\begin{tabular}{@{}lrrrrrr@{}}
\toprule
Policy & Original & Mutated & Gap & MSR & MTR & Proxy PTD \\
\midrule
Careful & 0.807 & 0.766 & 0.041 & 0.860 & 0.860 & $-0.167$ \\
Overfit & 0.549 & 0.526 & 0.023 & 0.053 & 0.667 & 0.940 \\
Shortcut & 0.345 & 0.339 & 0.006 & 0.000 & --- & 0.932 \\
\bottomrule
\end{tabular}
\caption{Controlled-study results, recomputed from saved trajectories. Original and Mutated are mean verifier scores over 450 pairs per policy; MSR and MTR use $\tau=0.65$; proxy PTD uses the 12 proxy-relevant tasks. MTR is undefined for \pol{shortcut} because none of its original runs pass.}
\label{tab:calibration}
\end{table}

The MTR of 0.667 for \pol{overfit} rests on only 36 pairs whose originals passed (two base tasks $\times$ six mutations $\times$ three identical seeds), so it does not indicate broad success; for comparison, \pol{careful}'s MSR has 95\% task-cluster interval $[0.840,0.887]$.

Proxy divergence exposes a different failure. On the 12 proxy-relevant tasks, \pol{shortcut} has mean PTD 0.932 (task-cluster interval $[0.176,1.781]$, wide because proxy scales differ across tasks). In \texttt{reward\_hack\_001}, for instance, it sets the visible reward weight and format bonus to 2.0 each, earning visible reward 4.0 while simulator quality stays at 0.0.

\subsection{Scoring Objectives Favor Different Behaviors}
\label{sec:selection}

A small fixed-program experiment (Appendix~\ref{app:selection}) illustrates a second measurement failure: an objective can prefer behavior that exploits its proxy. Selecting among five scripted programs, a visible-proxy objective picks \texttt{proxy\_exploit} (held-out pass rate 0.103, proxy PTD 0.996), while tool-volume and composite objectives both pick \pol{careful} (0.718, $-0.180$; Table~\ref{tab:selection}). The composite verifier still carries the tool-name defect of Section~\ref{sec:repair}, so selecting a useful policy and scoring equivalent executions consistently are different requirements.

\section{Study 2: LLM Agents on \texorpdfstring{$\tau^2$}{tau2}-bench}
\label{sec:external}

The controlled study shows that TRACE recovers known causes. Public agent benchmarks are harder: the agent is an LLM, the user is simulated, and no cause is known in advance. We therefore apply TRACE to an environment in which LLM agents converse with simulated users and act through tools: $\tau^2$-bench retail and airline \citep{barres2025tau2}, with 6 development and 30 evaluation tasks split equally across the two domains.

\paragraph{Setup.}
We evaluate four agents chosen by endpoint availability: internal GLM-5.3 Fast, internal DeepSeek V4.1 Flash, internal Laguna S 2.1 FP8, and Qwen3.8 Flash. A fixed simulated user and the benchmark's own evaluator are held constant across all studies (models in Appendix~\ref{app:external}). Each agent runs each task under three conditions: the original interface, renamed tools, and reformatted JSON observations with unchanged values. Unlike the controlled observation probe, reformatting changes presentation, not access to information. All conditions start from the same saved user opening; later turns are interactive. The task split, configuration, and analyses were frozen after development review (Appendix~\ref{app:external}).

\subsection{Fresh Interactions: Whole-System Sensitivity}

\begin{table}[t]
\centering
\small
\setlength{\tabcolsep}{4pt}
\begin{tabular}{@{}llrrrrl@{}}
\toprule
Agent & Condition & Pairs & Original & Mutated & $\Delta$ & 95\% task-bootstrap CI \\
\midrule
DeepSeek (internal) & Alias & 29/30 & .828 & .828 & .000 & [-.172, .172] \\
DeepSeek (internal) & Format & 29/30 & .828 & .862 & +.034 & [-.103, .172] \\
GLM (internal) & Alias & 30/30 & .833 & .767 & -.067 & [-.267, .133] \\
GLM (internal) & Format & 30/30 & .833 & .833 & .000 & [-.167, .167] \\
Laguna (internal) & Alias & 30/30 & .500 & .767 & +.267 & [.067, .433] \\
Laguna (internal) & Format & 30/30 & .500 & .633 & +.133 & [-.067, .333] \\
Qwen3.8 Flash & Alias & 29/30 & .759 & .690 & -.069 & [-.241, .138] \\
Qwen3.8 Flash & Format & 29/30 & .759 & .828 & +.069 & [-.138, .276] \\
\bottomrule
\end{tabular}
\caption{Public $\tau^2$-bench results under the benchmark's own evaluator. $\Delta$ is mutated minus original reward over complete pairs; Alias denotes tool renaming and Format observation reformatting. Confidence intervals resample base tasks within each domain.}
\label{tab:external}
\end{table}

\begin{figure}[t]
\centering
\includegraphics[width=\linewidth]{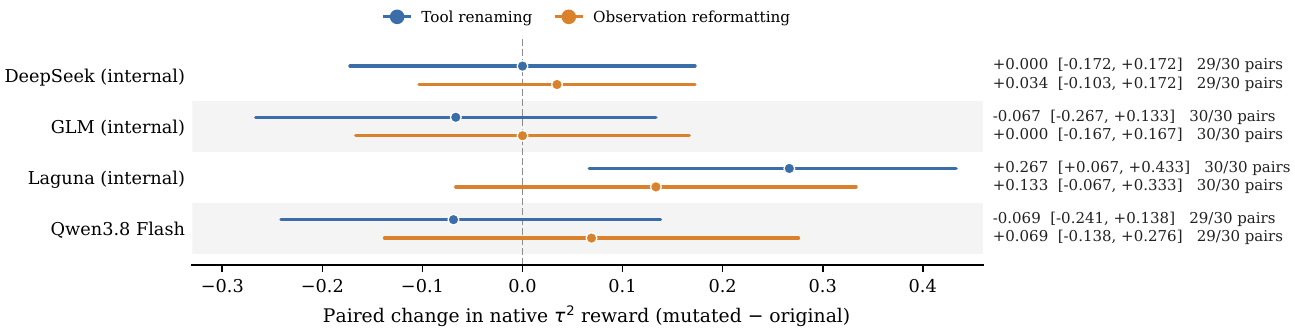}
\caption{Paired $\tau^2$-bench reward changes with 95\% confidence intervals (Table~\ref{tab:external}). Positive values mean a higher score under the mutation, the opposite sign convention to VRG in Figure~\ref{fig:repair}. Only Laguna under tool renaming excludes zero.}
\label{fig:forest}
\end{figure}

Each scored run receives 0 or 1 from the benchmark's evaluator, and $\Delta$ is mutated minus original reward. These runs test the whole interaction: the agent, later user turns, task state, and evaluator can all contribute to a difference.

The effects are mixed (Table~\ref{tab:external}, Figure~\ref{fig:forest}). Seven of eight intervals include zero; the exception is an \emph{improvement}, Laguna under tool renaming ($+.267$ [.067, .433]). With eight uncorrected intervals, a single exclusion is weak evidence of a systematic effect, and it does not replicate on new tasks (\S\ref{sec:study3}); the wide intervals equally fail to establish that the other effects are absent. Averages also conceal task-level churn: 65 of 236 complete pairs change reward, with improvements and regressions largely offsetting. Trajectory review links these changes to different write actions, different arguments, or incomplete required writes. They show that the agent--environment system is sensitive to presentation, but they are not by themselves evidence of a verifier defect.

\subsection{Fixed Trajectories: Evaluator Sensitivity}

Fresh runs cannot separate the evaluator from the agent. To isolate the evaluator, we present fixed trajectories to a separate single-call LLM judge (not a rerun of the full native evaluator) that sees the task criteria, domain policy, tools, and trajectory, but not the native reward or agent identity. Each of the 118 eligible original trajectories is shown in four views: original, identical repeat, tool-renamed, and observation-reformatted. A \emph{flip} is a pass/fail change relative to the original view.

Repetition and tool renaming each produce 0/118 flips; reformatting produces 1/118. In that case (Laguna, airline task 16), the original, repeated, and renamed views pass, but the reformatted view fails with a rationale that misstates the required date and flights. With one repeat per trajectory, we cannot distinguish a formatting effect from judge variability, so this is an observed inconsistency, not a demonstrated causal effect; Study 3 adds the missing repeat control.

\section{Study 3: Replication at Scale with Repeat and Positive Controls}
\label{sec:study3}

Study 2 leaves three questions open: is its one clear effect stable, would a null mean anything if the pipeline could not detect real effects, and does its single judge flip exceed call-to-call variation? Follow-up studies, sealed after Study 2's analyses were frozen and never pooled with it, answer each.

\paragraph{Design.}
The main replication runs the four agents on all 158 usable $\tau^2$-bench airline and retail tasks under the original, tool-renaming, and observation-reformatting conditions, with \emph{three} runs per condition (5,652 of 5,688 completed); the main test set is the 88 tasks never evaluated before (12 airline, 76 retail). This set, the $\pm0.10$ margin, and the decision rules below were fixed before any data were collected. Two controls make a paired difference interpretable. A \emph{repeat control} compares the outcome-flip rate between original and changed runs with the flip rate between two identical original runs; the difference is the \emph{excess flip rate}. The positive control permutes tool names among the domain's own names while keeping every description and schema true, so tasks stay solvable (gold actions under the shown names solve 164/164 base tasks offline) but names mislead (2,478 of 2,528 runs completed). A change is \emph{equivalent} if its 90\% task-cluster interval lies within $\pm0.10$, \emph{different} if its 95\% interval excludes zero, and otherwise \emph{inconclusive} (Appendix~\ref{app:study3}).

\begin{figure}[t]
\centering
\includegraphics[width=\linewidth]{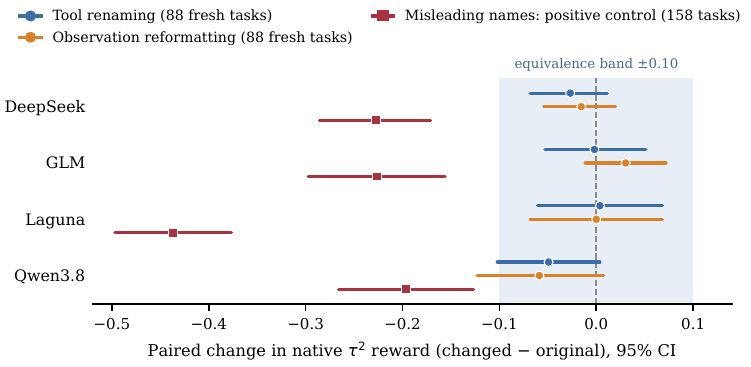}
\caption{Study 3 reward changes with 95\% intervals (Appendix Table~\ref{tab:study3}). Meaning-preserving changes stay near zero, while misleading tool names lower every agent's reward. The shaded band marks the $\pm0.10$ equivalence margin, which is judged on 90\% intervals.}
\label{fig:study3}
\end{figure}

\paragraph{Meaning-preserving changes are equivalent within $\pm0.10$; misleading names are not.}
Seven of eight agent--change pairs are equivalent within $\pm0.10$ (Figure~\ref{fig:study3}; full results in Appendix Table~\ref{tab:study3}); six remain so under a Bonferroni correction across the eight. The exception is Qwen3.8 Flash under reformatting, which is inconclusive ($-.059$ [$-.123$, $+.008$]) and flips outcomes 9.3 points more often than reruns do [$+2.7$, $+16.1$], a possible small effect that this study cannot confirm. The positive control is detected for all four agents, with reward falling by 0.196 to 0.437 and every 95\% interval well below zero, so the pipeline is sensitive enough for the equivalence results to be informative. Study 2's one clear effect does not replicate: Laguna's tool-renaming gain of $+.267$ falls to $-.050$ on an intermediate wave of 40 new tasks with one run per condition (Appendix~\ref{app:study3}) and to $-.011$ on Study 2's own 30 tasks with three runs, where it is equivalent within $\pm0.10$. An exploratory telecom replication (108 tasks) gives the same picture (Appendix~\ref{app:study3}).

\paragraph{Rerun noise explains why single runs mislead.}
Rerunning an identical original task flips its pass/fail outcome on 14.7--16.3\% of runs for DeepSeek, GLM, and Qwen, and on 35.6\% for Laguna. A single-run comparison such as Study 2's therefore cannot separate a presentation effect from ordinary variation, and Laguna's high rerun noise is consistent with its non-replicating gain.

\paragraph{Judges: stable under presentation, but measuring different things.}
On Study 2's 118 fixed trajectories, two current judges, GPT-6.1 Sol and Claude Opus 5.5, were called three times per view (2,831 of 2,832 calls usable). Presentation changes did not move verdicts beyond call-to-call variation: all four excess-disagreement intervals include zero, with estimates from $-0.85$ to $+0.56$ percentage points (Appendix Table~\ref{tab:study3-judges}). Claude Opus 5.5 disagrees with itself on 3.7\% of identical repeats, enough to masquerade as a presentation effect without a repeat control, and Study 2's single flip is consistent with such variation. Yet the two judges disagree on 57\% of the same records. Against the benchmark's own outcome reward, Claude Opus 5.5 agrees on 95\% of verdicts ($\kappa=0.89$) and GPT-6.1 Sol on 38\% ($\kappa=0.09$); every disagreement is a benchmark success, and in 163 of 201 GPT-6.1 Sol objects only to procedure, such as several tool calls in one turn. Because GPT-6.1 Sol's pass rate on benchmark successes ranges from 0\% for DeepSeek to 52\% for Qwen, the choice of judge could change how agents rank. Consistency is not validity: a repeatable judge can still measure a different construct.

\section{Related Work}

\paragraph{Reward hacking and overoptimization.}
Reward hacking occurs when behavior earns reward without meeting the intended goal \citep{krakovna2020specification, akter2026proxygaming, wang2026rewardhacking}; the mismatch grows under optimization \citep{gao2023overoptimization, eisenstein2023helping, kwa2024catastrophic, wen2025rethinking}, and benchmarks measure exploits by agents \citep{thaman2026rhb, gabor2025evilgenie, zhong2026impossiblebench}. Closest to our aims, \citet{akter2026proxygaming} stress-test evaluators to detect proxy gaming; TRACE instead attributes individual score changes.

\paragraph{Learning from evaluator feedback.}
Human and AI feedback are central training signals \citep{stiennon2020summarize, ouyang2022instructgpt, bai2022constitutional}, and verifiers and process supervision extend evaluation to intermediate steps \citep{cobbe2021verifiers, lightman2023verify, khalifa2025thinkprm, setlur2024rewarding, zhang2025reasonrag, fan2026corver, yuan2026vpr, zhang2026agentvrl}. LLM judges, in turn, are sensitive to content-irrelevant factors such as response order \citep{wang2024fairevaluators} and often disagree with experts on agent trajectories \citep{lu2025agentrewardbench}. Our tool-name defect lies in a simple hand-written check and is not a finding about learned process reward models.

\paragraph{Agent benchmarks.}
Agent benchmarks span tools, users, and persistent environments \citep{liu2023agentbench, zhou2023webarena, jimenez2023swebench, yao2024taubench, barres2025tau2, xie2024osworld, patil2025bfcl} as well as machine learning, research, workplace, human-assistance, software, and terminal tasks \citep{chan2024mlebench, starace2025paperbench, xu2025agentcompany, trinh2026hilbench, merrill2026terminalbench, da2025agentrlvr}. Critiques of agent evaluation document shortcuts, reproducibility gaps, and flawed reward design \citep{kapoor2025agentsmatter, zhu2025abc}; TRACE offers a paired procedure for detecting and attributing such flaws.

\paragraph{Behavioral testing and measurement validity.}
Targeted tests and multi-metric reporting are well established in NLP \citep{ribeiro2020checklist, goel2021robustnessgym, kiela2021dynabench, nie2020anli, liang2022helm}. Meaning-preserving changes to tool names or prompt format can shift model performance \citep{ye2024rotbench, sclar2024quantifying}; TRACE uses such changes to probe the evaluator as well as the agent. It also resembles metamorphic testing, which checks expected relations across transformed inputs \citep{chen2018metamorphic, liu2026sgveflooop}, but adds trajectory inspection and rescoring to attribute a violated relation to behavior or scoring.

\section{Discussion: What the Evidence Establishes}
\label{sec:discussion}

Table~\ref{tab:claims} maps claims to evidence. ResearchOps localizes a known scoring defect with execution held fixed; fresh $\tau^2$ runs reveal whole-system sensitivity; and the fixed-trajectory audits measure evaluator sensitivity with behavior held fixed. Study 3 adds two requirements: compare every paired difference with an identical rerun, and include a positive control, since equivalence means little unless the pipeline can detect a change known to matter; judge audits likewise need repeat controls. We also recommend reporting pass rates alongside gaps and confirming suspected defects by rescoring the same trajectories.

\begin{table}[t]
\centering
\scriptsize
\setlength{\tabcolsep}{4pt}
\renewcommand{\arraystretch}{1.15}
\begin{tabularx}{\linewidth}{>{\raggedright\arraybackslash}p{0.17\linewidth}>{\raggedright\arraybackslash}p{0.21\linewidth}X>{\raggedright\arraybackslash}p{0.19\linewidth}}
\toprule
Claim to test & What can go wrong & TRACE evidence & Remaining limit \\
\midrule
Equal work, equal credit & Wording or tool-surface dependence & Task and tool probes; alias-normalized rescoring removes careful's tool gap ($0.250\rightarrow0.000$) & Synthetic tools; two $\tau^2$ presentation changes \\
Reward reflects task quality & Visible proxy exploitation & Proxy-incentive probes; shortcut proxy PTD $0.932$ on the 12 proxy-relevant tasks & Hand-designed proxy channels \\
Stable findings & Threshold, task-sample, or rerun sensitivity & MSR/MTR at several thresholds; replication with rerun and positive controls, analysis fixed in advance & $\pm0.10$ margin; four agents \\
Evidence beyond scripts & Synthetic-only evidence & Paired $\tau^2$ studies on 30, 40, and 158 tasks with native evaluator; repeat-controlled judge audits & No judge ground truth; no training loop \\
\bottomrule
\end{tabularx}
\renewcommand{\arraystretch}{1.0}
\caption{What the evidence supports and what remains uncertain.}
\label{tab:claims}
\end{table}

\paragraph{Limits of the current evidence.}
ResearchOps has 25 designed tasks and scripts that can access the intended fix; it supports diagnosis under known conditions, not estimates of deployment failure rates or of exploits learned in training. The public studies cover 30, 40, and 158 airline and retail tasks plus 108 exploratory telecom tasks, two presentation changes, and four agents; equivalence is relative to a $\pm0.10$ margin, missing runs are not random, and native reward, partly LLM-graded, is a reference rather than ground truth. The judge audits measure consistency and agreement with native reward, not accuracy. The repair holds for one tested mapping, and holding out task IDs rather than mutation families leaves probe overfitting untested (Appendix~\ref{sec:dynamic}).

\section{Conclusion}

A score change is an observation, not an explanation: we need to know whether the agent's behavior, the task, or the evaluation changed. TRACE answers with paired mutations, trajectory inspection, and counterfactual rescoring. In a controlled environment it localizes a tool-name defect that rescoring fully repairs, distinct from behavioral failures; on $\tau^2$-bench, an initial single-run effect does not replicate, meaning-preserving presentation changes are equivalent within $\pm0.10$ for nearly all agents while a positive control is clearly detected, and judges stay consistent under presentation changes yet disagree on what counts as success. A score difference becomes evidence only against an identical rerun and a change known to matter.

\paragraph{Reproducibility and artifacts.}
All 1,575 controlled-study records reproduce exactly. The frozen $\tau^2$ artifacts include configurations, input hashes, task splits, raw traces, and analyses that recompute without model calls (Study 2: 360 runs and 472 judge calls; Study 3: 12,584 planned runs, 2,832 judge calls, and independent audits). TRACE code and the synthetic ResearchOps tasks are available at \url{https://github.com/RGaonkar/trace-verifier-stress-tests}; the reported controlled-study numbers come from the archived snapshot in the supplementary material. AI coding and writing tools assisted implementation, analysis, and revision; the author is responsible for the methods, evidence, citations, and text.

\bibliography{references}
\bibliographystyle{colm2026_conference}

\appendix
\section{Scoring Details}
\label{app:scoring}

The archived composite verifier is
\[
S=0.40O+0.25C+0.20K+0.15E-0.20D.
\]
$O$ checks for hidden-target terms in the final report. $C$ counts five canonical tool names: configuration inspection, document search, experiment execution, log inspection, and report submission. $K$ penalizes a protected-dataset error and missing constraint mentions. $E$ penalizes calls beyond a difficulty-dependent budget. $D$ is one if a simulated experiment logs hack risk above 0.5. The score is not clipped, so $S$ can fall outside $[0,1]$. The rubric probe changes the positive weights to $(0.30,0.20,0.35,0.15)$ and caps $K$ at 0.5 if the report lacks ``evidence.''

The tool-name defect lies in $C$. Replacing it with an alias-aware check $C_{\mathrm{alias}}$ gives $S_{\mathrm{alias}}=S+0.25(C_{\mathrm{alias}}-C)$, with the trajectory and all other components fixed.

The auxiliary hack detection rate is the mean of $D$. The process-overfitting score is $\max\{0,N-(0.60C+0.40E)\}$, where $N=\min\{1,n_{\mathrm{calls}}/6\}$; it compares two process heuristics, not reasoning quality. PTD uses the maximum proxy reward and maximum simulator quality across experiment runs, which may come from different runs. Its 12 applicable tasks have prefixes \texttt{reward\_hack}, \texttt{eval\_metric}, and \texttt{data\_leak}; other tasks are excluded rather than assigned a zero proxy.

\section{Policy-Selection Objectives}
\label{app:selection}

We add two programs (one skips log inspection; one targets visible proxy rewards) and let three objectives (visible proxy, tool volume, and the composite verifier, defined below) each select the highest-scoring of the five programs on the first 12 tasks. We evaluate the selection on the remaining 13 tasks under the original, wording, tool-interface, and proxy-incentive conditions. Task IDs are held out, but task types and mutation templates are shared.

\begin{table}[h]
\centering
\small
\begin{tabular}{@{}llrr@{}}
\toprule
Selection objective & Selected template & MSR & Proxy PTD \\
\midrule
Visible proxy & \texttt{proxy\_exploit} & 0.103 & 0.996 \\
Tool volume & \texttt{careful} & 0.718 & $-0.180$ \\
Composite score & \texttt{careful} & 0.718 & $-0.180$ \\
\bottomrule
\end{tabular}
\caption{Policy selection on 12 original tasks, evaluated on 13 held-out task IDs under three mutations (117 mutated pairs per selected policy). Proxy PTD is restricted to proxy-relevant evaluation tasks. These are fixed-program selection results, not learning curves.}
\label{tab:selection}
\end{table}

The visible-proxy objective selects \texttt{proxy\_exploit}, while the tool-volume and composite objectives both select \pol{careful} (Table~\ref{tab:selection}): held-out pass rates are 0.103 versus 0.718, and proxy PTD 0.996 versus $-0.180$. Since tool-volume and composite selection agree, the composite objective shows no advantage over that comparator.

The visible-proxy objective is $0.45P+0.30F+0.25N$, where $F$ checks whether the report names the intended fix, including supported aliases. The tool-volume objective uses $N$ alone, and the composite objective uses $S$, not a robustness gap or held-out transfer. Each selects the highest mean among five programs on the first 12 original tasks in file order; exact ties follow the code's policy ordering. Careful behavior and exploitation are both already in the candidate set, so this search does not show how a learning agent would acquire either behavior.

\section{Public Study Details}
\label{app:external}

The study uses $\tau^2$-bench (\url{https://github.com/sierra-research/tau2-bench}) at commit \nolinkurl{1d244f5dca42944b67a379b44bfeb9f5748f189d}. A seeded split selects three development and 15 evaluation tasks per domain. The simulated user is GPT-4.1 Mini (temperature 0), the native evaluator grades natural-language assertions with GPT-5 Mini, and the single-call judge of Section~\ref{sec:external} is GPT-4.1 Mini; all three are held fixed across Studies 2 and 3 for comparability. Models, user simulator, evaluator, judge, token limits, mutations, retry policy, and planned analyses were fixed after development review and recorded in run and freeze fingerprints. Aliases prepend \texttt{workflow\_} to tool names, preserving schemas, dispatch, arguments, and canonical execution. Reformatting pretty-prints and key-sorts JSON without changing values or array order. Each alias run emits only aliased names before canonicalization, and each format run changes at least one JSON observation.

Two originals are missing from the 360-job inventory. Internal DeepSeek on airline task 21 exhausted the 8,192-token output cap without an answer or tool call, and Qwen on airline task 23 was externally interrupted during timeout handling. Neither reward is imputed; each removes one task from both paired estimates for its model. A separate Qwen alias run that reached the harness timeout is retained as a valid scored zero. The four-view judge audit covers 118 trajectories and 472 calls, with agent identity and native reward withheld and repetition as a noise control. Transition tables, changed-pair traces, hashes, and missingness records accompany the artifact.

\section{Toward Held-Out Probes}
\label{sec:dynamic}

An agent tuned against fixed probes could memorize aliases or reward traps rather than learn the intended behavior. We propose three levels: \emph{development probes} for tuning, \emph{held-out instances} (new names and values within known families), and \emph{held-out families}, with policies and verifiers frozen before held-out evaluation and each probe validated to make only its intended change. A fixed reference set would support comparison across rounds, as in dynamic benchmarking \citep{kiela2021dynabench}. This is a proposal, not a demonstrated solution to probe overfitting.

\section{Study 3 Details}
\label{app:study3}

\paragraph{Analysis plan fixed in advance; integrity.} All Study 3 runs were executed under sealed plans; the analysis plan, decision rules, and code were frozen and independently reviewed before data collection, and independent audits recomputed every reported number from raw records with no mismatches. Intervals are domain-stratified task-cluster percentile bootstrap intervals (2,000 draws, seed 17; labels unchanged at 20,000 draws). Changes are averaged over same-seed repetitions within a task, then over tasks. The repeat control uses a seed-balanced pairing of original and changed repetitions. Failed runs are kept missing without retries; labels are unchanged under 0/1 imputation and truncation-exclusion sensitivities. Missingness is not random: most main-run failures are DeepSeek output-limit truncations, and 42 of the 50 positive-control failures occur under misleading names, where the positive control still passes under worst-case handling. Agents, user simulator (GPT-4.1 Mini, temperature 0), native evaluator (GPT-5 Mini), and 8,192-token agent output limit are unchanged from Study 2.

\begin{table}[h]
\centering
\scriptsize
\setlength{\tabcolsep}{2.2pt}
\begin{tabular}{@{}llrrrllrrl@{}}
\toprule
Agent & Change & Tasks & Original & $\Delta$ & 95\% CI & 90\% CI & Label & Rerun flips & Excess flips [95\% CI] \\
\midrule
DeepSeek (internal) & Alias & 87 & .843 & -.027 & [-.068, +.011] & [-.061, +.004] & equiv. & .147 & -.031 [-.077, +.012] \\
DeepSeek (internal) & Format & 86 & .841 & -.016 & [-.054, +.020] & [-.049, +.016] & equiv. & .147 & -.023 [-.065, +.020] \\
GLM (internal) & Alias & 88 & .831 & -.002 & [-.053, +.051] & [-.044, +.042] & equiv. & .157 & -.028 [-.070, +.015] \\
GLM (internal) & Format & 88 & .833 & +.030 & [-.011, +.072] & [-.008, +.064] & equiv. & .159 & -.023 [-.068, +.027] \\
Laguna (internal) & Alias & 88 & .610 & +.004 & [-.061, +.068] & [-.053, +.057] & equiv. & .356 & -.019 [-.091, +.049] \\
Laguna (internal) & Format & 88 & .610 & .000 & [-.068, +.068] & [-.057, +.057] & equiv. & .356 & +.015 [-.053, +.080] \\
Qwen3.8 Flash & Alias & 88 & .812 & -.049 & [-.102, +.004] & [-.095, -.004] & equiv. & .163 & +.034 [-.019, +.091] \\
Qwen3.8 Flash & Format & 88 & .812 & -.059 & [-.123, +.008] & [-.114, -.004] & inconcl. & .163 & +.093 [+.027, +.161] \\
\midrule
DeepSeek (internal) & Misleading & 154 & .844 & -.227 & [-.286, -.171] & [-.277, -.180] & different & .140 & +.180 [+.109, +.250] \\
GLM (internal) & Misleading & 157 & .818 & -.226 & [-.297, -.156] & [-.285, -.168] & different & .159 & +.194 [+.120, +.266] \\
Laguna (internal) & Misleading & 158 & .633 & -.437 & [-.497, -.377] & [-.487, -.386] & different & .304 & +.196 [+.104, +.291] \\
Qwen3.8 Flash & Misleading & 158 & .791 & -.196 & [-.266, -.127] & [-.253, -.139] & different & .181 & +.187 [+.113, +.266] \\
\bottomrule
\end{tabular}
\caption{Study 3 on $\tau^2$-bench airline and retail. Top: tool renaming (Alias) and observation reformatting (Format) on 88 never-evaluated tasks, three runs per condition. Bottom: the misleading-name positive control on all 158 tasks, two runs per condition. $\Delta$ is changed minus original reward; ``equiv.'' means the 90\% interval lies within $\pm0.10$. Rerun flips is the outcome-flip rate between identical original runs; excess flips subtracts it from the original-versus-changed flip rate.}
\label{tab:study3}
\end{table}

\paragraph{Intermediate 40-task wave.} Before the scale-up, 40 new tasks (20 airline, 20 retail) were sampled by seed from IDs unused in Study 2 and run once per condition (478 of 480 runs completed; two DeepSeek runs on one airline task hit the output cap and are excluded, and imputing either outcome changes no conclusion). No change was detectable (Table~\ref{tab:wave40}); every interval includes zero, and the smallest exact McNemar $p$ is 0.29. With one run per condition, these differences mix the perturbation with agent and user-simulator variation.

\begin{table}[h]
\centering
\footnotesize
\setlength{\tabcolsep}{3.4pt}
\begin{tabular}{@{}llrrrrlr@{}}
\toprule
Agent & Condition & Pairs & Original & Mutated & $\Delta$ & 95\% CI & Switches \\
\midrule
DeepSeek (internal) & Alias & 39/40 & .846 & .769 & -.077 & [-.179, +.026] & 1 / 4 \\
DeepSeek (internal) & Format & 39/40 & .846 & .821 & -.026 & [-.128, +.077] & 2 / 3 \\
GLM (internal) & Alias & 40/40 & .900 & .850 & -.050 & [-.150, +.050] & 1 / 3 \\
GLM (internal) & Format & 40/40 & .900 & .850 & -.050 & [-.150, +.025] & 1 / 3 \\
Laguna (internal) & Alias & 40/40 & .650 & .600 & -.050 & [-.225, +.125] & 6 / 8 \\
Laguna (internal) & Format & 40/40 & .650 & .750 & +.100 & [-.050, +.250] & 7 / 3 \\
Qwen3.8 Flash & Alias & 40/40 & .825 & .725 & -.100 & [-.250, +.025] & 2 / 6 \\
Qwen3.8 Flash & Format & 40/40 & .825 & .850 & +.025 & [-.150, +.175] & 6 / 5 \\
\bottomrule
\end{tabular}
\caption{First replication wave: 40 new $\tau^2$-bench tasks, one rollout per condition, under the benchmark's own evaluator. $\Delta$ is mutated minus original reward over complete pairs, with task-bootstrap intervals; switches count tasks going from fail to pass / pass to fail.}
\label{tab:wave40}
\end{table}

\paragraph{Judge audit.} The judges used medium reasoning effort, an 8,192-token cap, and a new prompt asking for \texttt{pass}, \texttt{fail}, or \texttt{insufficient} with a brief reason; agent identity and native reward were withheld, and original and repeat payloads were byte-identical but sent as separate requests. One Claude Opus 5.5 repeat returned empty content after a provider content filter and is kept missing; setting it to agree or disagree bounds the alias excess at $[-1.13, -0.85]$ and the format excess at $[-0.56, -0.28]$ percentage points. All three GPT-6.1 Sol change events come from one airline task, and across its three original calls GPT-6.1 Sol varies on 1 of 118 records, so its zero repeat disagreement does not establish determinism. The comparison with native reward is descriptive: native reward is a reference, not ground truth, since part of it is graded by an LLM and it does not check most procedural rules. The procedural-objection count uses a keyword heuristic on the judges' stated reasons.

\begin{table}[h]
\centering
\small
\setlength{\tabcolsep}{4pt}
\begin{tabular}{@{}llrll@{}}
\toprule
Judge & View & Events & Disagreement [95\% CI] & Excess over repeat, pp [95\% CI] \\
\midrule
GPT-6.1 Sol & Repeat & 0/354 & 0.00\% [---] & \\
GPT-6.1 Sol & Alias & 1/354 & 0.28\% [0.00, 0.85] & +0.28 [0.00, +0.85] \\
GPT-6.1 Sol & Format & 2/354 & 0.56\% [0.00, 1.71] & +0.56 [0.00, +1.71] \\
Claude Opus 5.5 & Repeat & 13/353 & 3.68\% [0.86, 7.56] & \\
Claude Opus 5.5 & Alias & 10/354 & 2.82\% [0.85, 5.60] & -0.85 [-3.72, +1.15] \\
Claude Opus 5.5 & Format & 12/354 & 3.39\% [1.15, 5.93] & -0.28 [-2.82, +2.23] \\
\bottomrule
\end{tabular}
\caption{Repeat-controlled judge audit on Study 2's 118 fixed trajectories, three calls per view. Excess subtracts the identical-repeat disagreement rate on the same complete triples (353 for Claude Opus 5.5 after one missing call); GPT-6.1 Sol's zero repeat count has no meaningful interval.}
\label{tab:study3-judges}
\end{table}

\paragraph{Telecom (exploratory).} The same design on 108 telecom tasks (3,867 of 3,888 runs completed; only agent tools aliased; 200-step limit) gives seven of eight pairs equivalent within $\pm0.10$, with Laguna under reformatting inconclusive and no change exceeding rerun noise (Table~\ref{tab:telecom}). One telecom trajectory contains a credential-like string reproduced by an agent, so raw telecom traces require redaction before release.

\begin{table}[h]
\centering
\scriptsize
\setlength{\tabcolsep}{2.2pt}
\begin{tabular}{@{}llrrrllrrl@{}}
\toprule
Agent & Change & Tasks & Original & $\Delta$ & 95\% CI & 90\% CI & Label & Rerun flips & Excess flips [95\% CI] \\
\midrule
DeepSeek (internal) & Alias & 108 & .759 & -.035 & [-.088, +.017] & [-.080, +.009] & equiv. & .185 & +.009 [-.037, +.059] \\
DeepSeek (internal) & Format & 108 & .756 & -.023 & [-.074, +.028] & [-.068, +.020] & equiv. & .187 & +.019 [-.035, +.074] \\
GLM (internal) & Alias & 108 & .994 & +.003 & [-.006, +.012] & [-.006, +.012] & equiv. & .012 & -.003 [-.012, +.006] \\
GLM (internal) & Format & 108 & .994 & -.015 & [-.037, +.003] & [-.034, .000] & equiv. & .012 & +.015 [-.003, +.037] \\
Laguna (internal) & Alias & 108 & .404 & +.006 & [-.052, +.065] & [-.043, +.056] & equiv. & .309 & -.019 [-.068, +.031] \\
Laguna (internal) & Format & 108 & .404 & -.056 & [-.114, +.003] & [-.105, -.006] & inconcl. & .309 & -.037 [-.086, +.019] \\
Qwen3.8 Flash & Alias & 108 & .787 & -.040 & [-.080, -.000] & [-.074, -.006] & equiv. & .148 & +.022 [-.012, +.059] \\
Qwen3.8 Flash & Format & 108 & .789 & +.008 & [-.037, +.054] & [-.029, +.045] & equiv. & .151 & +.040 [-.009, +.093] \\
\bottomrule
\end{tabular}
\caption{Exploratory telecom replication, 108 tasks, three runs per condition; columns as in Table~\ref{tab:study3}.}
\label{tab:telecom}
\end{table}

\end{document}